\documentclass[letterpaper, 10 pt, journal, twoside]{IEEEtran}

\ifCLASSINFOpdf

\else

\fi

\usepackage{amsmath}

\usepackage{colortbl} % Add this to your preamble
\usepackage[table,xcdraw]{xcolor} % For color support
\usepackage{xcolor}
\usepackage{graphicx}
\usepackage{gensymb}
\usepackage{hhline}
\usepackage{url}
\usepackage{cite}
\usepackage{float}
\usepackage{caption}
\usepackage[colorlinks,allcolors=blue]{hyperref}

\begin{document}
% \IEEEoverridecommandlockouts
\title{Neuromorphic Attitude Estimation and Control}
\author{S. Stroobants, C. De Wagter, and G.\,C.\,H.\,E. de Croon%
        % <-this % stops a space
\thanks{Manuscript received: Nov, 20, 2024; Revised Feb, 13, 2025; Accepted Mar, 14, 2025.}
\thanks{This paper was recommended for publication by Editor Tetsuya Ogata upon evaluation of the Associate Editor and Reviewers' comments.}% <-this % stops a space
\thanks{This material is based upon work supported by the Air Force Office of Scientific Research under award number FA8655-20-1-7044.}% <-this % stops a space
\thanks{All authors are with the Micro Air Vehicle Laboratory, Faculty of Aerospace Engineering, Delft University of Technology, The Netherlands }% <-this % stops a space
\thanks{Digital Object Identifier (DOI): see top of this page.}}
\maketitle

\markboth{IEEE Robotics and Automation Letters. Preprint Version. Accepted March, 2025}
{Stroobants \MakeLowercase{\textit{et al.}}: Neuromorphic Attitude Estimation and Control}

\begin{abstract}
The real-world application of small drones is mostly hampered by energy limitations. Neuromorphic computing promises extremely energy-efficient AI for autonomous flight but is still challenging to train and deploy on real robots. To reap the maximal benefits from neuromorphic computing, it is necessary to perform all autonomy functions end-to-end on a single neuromorphic chip, from low-level attitude control to high-level navigation. This research presents the first neuromorphic control system using a spiking neural network (SNN) to effectively map a drone's raw sensory input directly to motor commands. We apply this method to low-level attitude estimation and control for a quadrotor, deploying the SNN on a tiny Crazyflie. We propose a modular SNN, separately training and then merging estimation and control sub-networks. The SNN is trained with imitation learning, using a flight dataset of sensory-motor pairs. Post-training, the network is deployed on the Crazyflie, issuing control commands from sensor inputs at 500Hz. Furthermore, for the training procedure we augmented training data by flying a controller with additional excitation and time-shifting the target data to enhance the predictive capabilities of the SNN. On the real drone, the perception-to-control SNN tracks attitude commands with an average error of 3.0 degrees, compared to 2.7 degrees for the regular flight stack. We also show the benefits of the proposed learning modifications for reducing the average tracking error and reducing oscillations. Our work shows the feasibility of performing neuromorphic end-to-end control, laying the basis for highly energy-efficient and low-latency neuromorphic autopilots.
\end{abstract}


% \IEEEpeerreviewmaketitle
\vspace{-10pt}
\section{Introduction}
\IEEEPARstart{Q}{uadrotors} have soared in popularity over the past decade, significantly influencing the field of unmanned aerial vehicles (UAVs) with their unique capabilities. These agile machines are applicable in a myriad of applications, such as search and rescue operations~\cite{daud2022applications}, environmental monitoring~\cite{tang2015drone} and precision agriculture~\cite{mogili2018review}, owing to their ability to hover, perform vertical take-offs and landings, and navigate through confined spaces with remarkable precision.

The integration of Artificial Intelligence (AI) promises to extend the capabilities of quadrotors even further~\cite{song2023reaching,azar2021drone}. By leveraging advances in AI, we can envision quadrotors that not only perform pre-programmed tasks but also adapt to new challenges, achieving levels of flight performance and operational robustness previously unattainable while solving tasks that are currently performed post-flight or offboard. However, the current generation of quadrotors is hindered by hardware that is often power-hungry and algorithms that fall short in efficiency and adaptability~\cite{sunderhauf2018limits}.

A promising solution to these challenges lies in the emerging field of neuromorphic hardware~\cite{sandamirskaya2022neuromorphic}. Neuromorphic systems, including processors and sensors such as event-based cameras~\cite{gallego2020event,lichtsteiner2008128}, draw inspiration from neural systems found in nature. These systems use sparse and asynchronous spikes to transmit information that are both energy-efficient and enable high-speed processing. Due to the low latency, this approach is particularly well-suited for dynamic environments where rapid decision-making is crucial~\cite{indiveri2000neuromorphic}.
\begin{figure}[t!]
\centering
\includegraphics[width=0.48\textwidth]{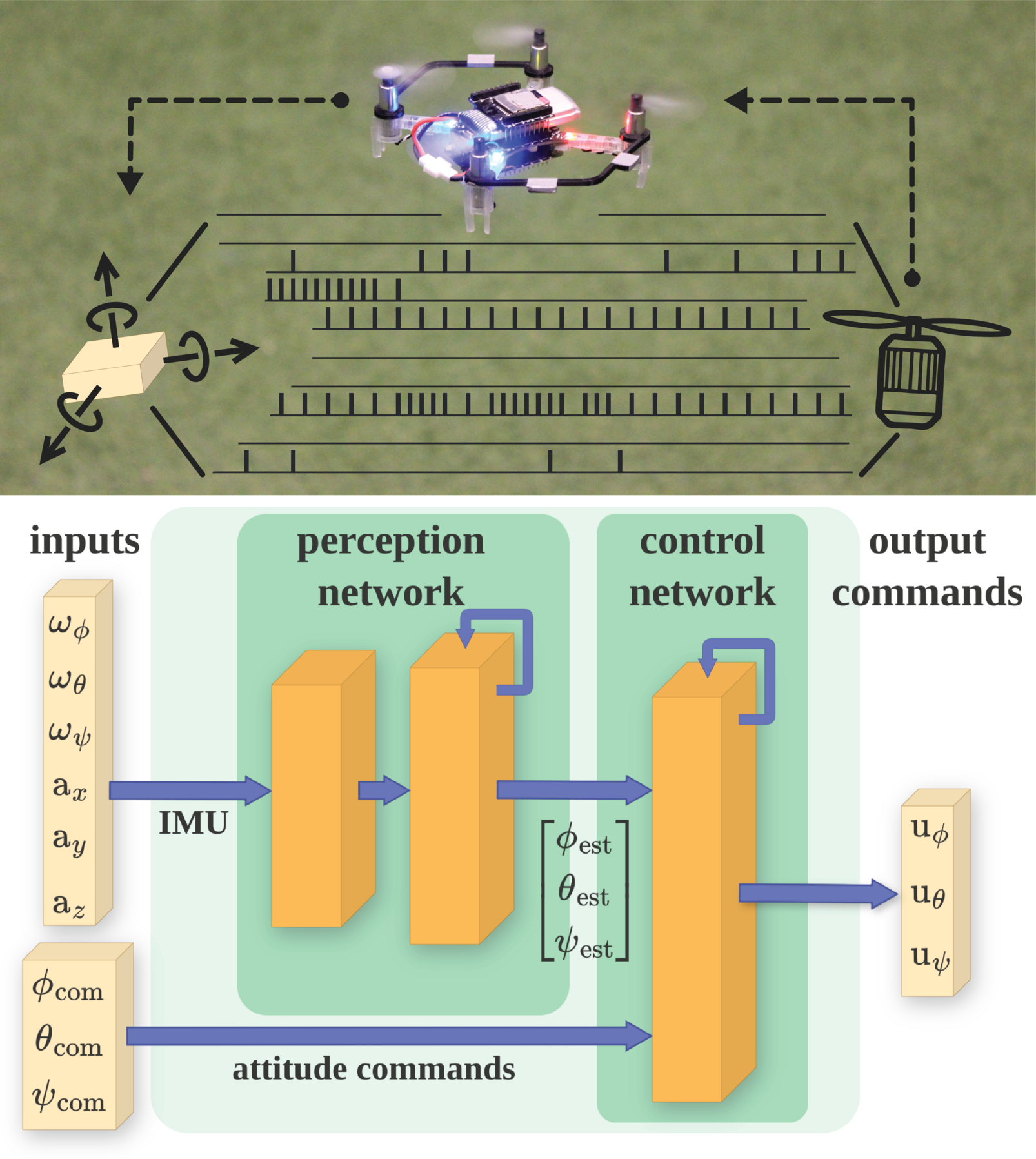}
\caption{We present an approach to training a spiking neural network for end-to-end attitude estimation and control of tiny drones (deployed on a Crazyflie, top). The network is a merging of a 2-layer attitude estimation sub-network with recurrency and a 1-layer recurrent attitude control network (bottom). The network exhibits a spiking activity of $15\%$, which is promising in terms of energy efficiency for future implementation on a neuromorphic processor. The network currently runs at 500Hz on a Teensy microcontroller.}
\vspace{-15pt}
\label{fig:fig1}
\end{figure}
Central to this neuromorphic paradigm are Spiking Neural Networks (SNNs)~\cite{maass1997networks}, which emulate the brain’s information processing using neural spikes. SNNs have demonstrated their potential in various robotic applications, yet their use in controlling the full flight dynamics of quadrotors remains largely unexplored. 
By adopting strategies seen in nature, such as the reflexive control and visual processing used by the fruit fly~\cite{dickinson1999haltere}, we can develop more integrated and efficient control systems. This does, however, require a fully end-to-end neuromorphic system.

Neuromorphic control is a nascent field at the intersection of neuroscience and robotics control theory~\cite{bartolozzi2022embodied}. 
The benefits of neuromorphic hardware, such as fast inference and high energy efficiency~\cite{schuman2017survey}, harmonize with demanding control and estimation tasks. 
While the output of rudimentary sensors for quadrotors, such as Inertial-Measurement-Units, can already be processed at the high frequencies necessary for agile and robust control, vision-based tasks are severely limited by processing power on a flying machine~\cite{pellerito2023end}. 
However, Dimitrova~\textit{et al.}~\cite{dimitrova2020towards} have shown that using event-based cameras allows a quadrotor to track the horizon at extremely high speeds. To further increase the potential of such a system, the authors of~\cite{vitale2021event} showed that integration of this horizon tracker with a manually-tuned SNN controller on a single neuromorphic processor leads to even faster control, benefiting from having all parts on the same chip. 

Despite significant advances in AI for quadrotors, limitations remain, particularly in vision-based tasks constrained by onboard computational resources. 
Falanga~\textit{et al.}~\cite{falanga2020dynamic} argue that regular frame-based cameras are inadequate for avoiding obstacles due to their high latency, which can be detrimental in fast-paced environments. 
Although event-based cameras address these latency issues, the processing on non-neuromorphic hardware required compromises in detection algorithms to favor speed over accuracy.

Recent breakthroughs in quadrotor research have achieved impressive results, such as outperforming human pilots in drone races using only onboard computations~\cite{kaufmann2023champion}. Also, Song \textit{et al.}~\cite{song2023reaching} show that for these tasks, optimal control methods are no longer sufficient and are beaten by Reinforcement Learning (RL) employing Deep Learning. 

Despite these accomplishments, the reliance on slower frame-based vision systems, typically operating at 30Hz or lower, highlights a significant gap where neuromorphic solutions could offer substantial improvements. These examples underscore the critical need for fully integrated neuromorphic systems capable of high-speed data processing.

To allow such a unified system, the entire estimation and control loop needs to be considered. 
Despite the promising results in partial implementations, a fully integrated neuromorphic system connecting sensor inputs directly to motor commands has not yet been realized in operational quadrotors.
Results focusing exclusively on lower-level SNN control have been obtained using manually tuned networks~\cite{stagsted2020towards, stroobants2022design} or were limited to simulation~\cite{clawson2016spiking, qiu2020evolving}. Moreover, even state-of-the-art learned quadrotor controllers using regular Multilayer Perceptrons (MLPs) as presented in~\cite{hwangbo2017control, kaufmann2020deep} and~\cite{ferede2024end}, that were learned with RL, assume full state knowledge or need a lower-level controller to go from rate commands to motor outputs. 
Zhang~\textit{et al.}~\cite{zhang2016learning} have demonstrated in simulation that by using an expert privileged policy, an MLP can be trained to perform end-to-end control. But also here the observation model, containing the measurements, included a direct measurement of the drone's attitude. 
However, such privileged information -- complete and accurate state information -- is rarely available in real-world scenarios. 
This limitation is further exacerbated by the reality gap, that arises when algorithms trained or evaluated in simulation must cope with real-world conditions characterized by imperfect measurements, sensor noise, actuator delays, and unpredictable environmental influences.

Notable efforts towards a complete end-to-end neuromorphic system include the use of Intel's Loihi processor~\cite{davies2018loihi} in a quadrotor for velocity control based on optical flow estimates from event-based cameras~\cite{paredes2023fully}, which successfully combined ego-motion estimation with a basic linear controller. 
The experimental results confirmed the potential of neuromorphic technology, as the vision ran at frequencies between 274-1600Hz, while only spending 7mW for network inference compared to 14-25Hz on a Jetson Nano that required 1-2W for inference. 
The neuromorphic system was not only significantly faster, but also required orders of magnitude less power. 
However, it still relied on a companion computer for attitude control, introducing delays, increasing power consumption, and adding weight to the drone.
Moreover, the linear neuromorphic controller lacked a mechanism to compensate for steady-state errors, such as those caused by sensor biases like gyroscope drift.
With our work, we want to demonstrate how the pipeline of \cite{paredes2023fully} could be extended to run on a single neuromorphic chip. 
In~\cite{slijkhuis2023closed} a closed-form spiking network was proposed that could do end-to-end control and estimation for linear systems and was shown to perform well with a small number of neurons in simulation. Since this approach needs to be able to read out a floating point "firing rate" of neurons in the hidden layer, it is not trivially implemented on commonly available neuromorphic hardware where the input and outputs are limited to vectors of binary spikes. 

The \textbf{main contribution} of this article is that we design, train, and implement the first fully neuromorphic system for attitude estimation and control of quadrotors. 
The proposed method involves real-time processing from sensors to actuators and does not require traditional computing hardware. 
Our approach is to train two separate sub-networks, one for state estimation and one for control, and to merge them after training. For both parts of the network, we employ supervised / imitation learning. 
In our creation of the training scheme we had to overcome substantial challenges, as the spiking neural network needs to cope with (i) sensor bias, (ii) delays due to the progressive updates of spiking neural networks, (iii) the reality gap and (iv) converting binary spikes to a motor command that leads to smooth control. 
Additional contributions of our work concern how we tackled these challenges. 
For the sensor bias, we find that constraining the parameters of a small subgroup of neurons to function as integrators is necessary for successful training results. 
These integrator neurons can now operate analogously to the integral component of a standard PID controller, effectively mitigating persistent sensor biases. 
For the delays in the SNN, we propose to time-shift the targets for learning, so that the SNN predicts future outputs of the traditional controller. This brings substantial performance improvement. For the reality gap, we first add noise to the motor outputs of the traditional controller to sufficiently excite the system and avoid biases in learning. Subsequently, we gather more training data with a first version of the SNN, so that relevant off-target attitudes and rates are explored. Finally, we evaluate system performance in real-world conditions, comparing the trained SNN with traditional control methods. 

The remainder of the article is structured as follows. Section~\ref{sec:methodology} details our methodology, covering attitude control from sensor data, the network architecture, training procedures, and the hardware used for real-world testing. In Section~\ref{sec:results}, we present the test results, including position control, attitude control, and an analysis of power consumption. Finally, Section~\ref{sec:conclusion} summarizes our key findings and outlines potential directions for future work in neuromorphic control systems.

\section{Methodology}
\label{sec:methodology}
This section discusses how an SNN used for attitude estimation and control of the Crazyflie in real time, was constructed and trained. 
\vspace{-8pt}
\subsection{Attitude Control from IMU measurements}
The attitude of a quadrotor, its orientation relative to gravity, can be estimated using measurements from an Inertial Measurement Unit (IMU). 
These IMUs commonly contain a 3 DOF (Degree of Freedom) gyroscope, measuring rotational velocities and a 3 DOF accelerometer, measuring linear acceleration.
The gyroscope data offers high-frequency information about the rotation of the quadrotor while the accelerometer measurements contain an absolute measurement of the gravity vector~\cite{martin2010true}. 
Combined, these two form the backbone of most quadrotor control algorithms. 
These 6 inputs are usually combined into an estimate of the orientation of the drone, which in turn gets sent to a controller together with a target orientation. This controller calculates the necessary motor speeds for each four rotors. 
\vspace{-8pt}
\subsection{Spiking Neural Network Architecture}
\subsubsection{LIF neurons}
In this work, we apply one of the most common spiking neuron models; the current-based leaky-integrate-and-fire (CUBA-LIF) neuron. 
This model is chosen since it captures temporal dynamics, is computationally efficient and is the default model in current available neuromorphic platforms such as Intel's Loihi~\cite{davies2018loihi}. Each neuron is connected to other neurons via \textit{synapses}, connections that carry a multiplicative weight. 
Every neuron keeps track of two hidden states at each timestep; its \textit{membrane potential} and \textit{synaptic current}. The membrane potential $\upsilon$ and synaptic input current $i$ at timestep $t$ as discrete functions of time are given as: 
\begin{align}
    \label{eq:LIF}
    \upsilon_i(t+1) &= \tau^{\text{mem}}_ i\upsilon_i(t) + i_i(t), \\
    i_i(t+1) &= \tau^{\text{syn}}_i i_i(t) + \sum w_{ij} s_j(t) + \sum w_{ik} s_k(t),
\end{align}
where $j$ and $i$ denote presynaptic (input) and postsynaptic (output) neurons within a layer, $k$ the neurons in the same layer as $i$, $s \in [0, 1]$ a neuron spike, and $w^{ij}$ and $w^{ik}$ feedforward and recurrent connections (if any), respectively. The leak values of the two internal state variables are denoted by $\tau_i^{\text{mem}}$ and $\tau_i^{\text{syn}}$. A neuron fires an output spike if the membrane potential $\upsilon_i$ exceeds threshold $\theta_i$ to all connected neurons, resetting its membrane potential to zero at the same time. 

The input of the networks during training is a linear layer that is directly inserted into the current $i$ of the first layer. 
This way, the encoding of floating point sensor data to binary spikes is included in the training procedure.
The output is decoded similarly; the hidden spiking layer is connected via a weight matrix to the outputs.  

\subsubsection{Combination of networks}
To facilitate learning of specific tasks and increase the debugability, the training is split into two parts; estimation and control. 
By learning layers of spiking neurons that have a certain function, there is more control over the stability of the final solution, and it also reduces the search space.
Since we define the input- and output values of both sub-networks as a linear multiplication of the input- or output-vector respectively, the networks can be easily combined. 
The output of the first network can be written as $y(t) = W_\text{o} s(t)$, with $s(t)$ the spikes in the hidden layer, and the input to the next network is $x(t) = W_\text{i} y(t)$. 
We can now combine these by multiplying the weight matrices of the output weights $W_\text{o}$ of the first network and the input weights $W_\text{i}$ of the second, as introduced in~\cite{paredes2023fully}, since these are both linear transformations. The attitude part of the input to the second network can therefore be written as
\begin{equation}
    \begin{bmatrix}
        \phi_\text{est} \\
        \theta_\text{est} \\
        \psi_\text{est}
    \end{bmatrix} = W_\text{i} W_\text{o} s(t).
\end{equation}
Stacking the binary output spikes of the first network with the floating-point command values that are passed (see Figure~\ref{fig:fig1}), the new set of weights to the hidden layer of the second network can be written as
\begin{equation}
    W_\text{new} = 
    \begin{bmatrix}
        0 & W_\text{i, command} \\
        W_\text{i} W_\text{o} & 0 \\
    \end{bmatrix}.
\end{equation}
\vspace{-8pt}
\subsection{Training}
\label{subsec:training}
The model is trained using imitation learning, cloning the behavior of an expert policy. 
Data is gathered at 500Hz by flying manually with a Crazyflie for 20 minutes.
During these tests, the Crazyflie uses a complementary filter for estimating the attitude and a cascaded PID controller for control. In this work, these function as the expert policy.
The Crazyflie controller used the default parameters as defined by the Bitcraze firmware~\cite{giernacki2017crazyflie}.
This data was split into sequences of 2000 timesteps and normalized according to total training set statistics. 
From every sequence the integrator value at the beginning of this sequence was subtracted, since this value is not contained in the input data so would not be possible to learn.
All of the parameters $p$ of the network ($\tau_i^\text{mem}$, $\tau_i^\text{syn}$, $w_{ij}$, $w_{ik}$ and $\theta_i$) were then trained using supervised backpropagation-through-time (BPTT). 
The loss was defined as a weighted sum of the Mean Squared Error (MSE) and the Pearson Correlation Loss; 
\begin{equation}
    J(p) = \text{MSE}(x, \hat{x}) + \frac{1}{2}(1 - \rho(x, \hat{x})),
\end{equation}
with $x$ and $\hat{x}$ the target- and network response values respectively and $\rho(x, \hat{x})$ the Pearson Coefficient~\cite{cohen2009pearson}.
One major step in training SNNs using regular BPTT despite the non-differentiability of the spiking threshold function is replacing the Heaviside step-function in the backwards pass with a surrogate function that represents a smooth approximation of the real gradient~\cite{neftci2019surrogate}. In this work, the derivative of a scaled arctangent was used, like in~\cite{fang2021incorporating};
\begin{equation} \label{eq:surrogate}
    \frac{d}{dx} \left( \frac{1}{s} \arctan(s x) \right) = \frac{1}{1 + (s x)^2},
\end{equation}
where $s$ is the slope of the surrogate. A higher slope results in a more accurate proxy of the real gradient, but can lead to vanishing gradients for neurons with a very low or high membrane potential. A shallow slope, on the other hand, is less accurate but leads to less "dead" neurons that have no contribution to the output. Among alternatives for the surrogate gradient is the derivative of the Sigmoid, but research has shown that the exact shape does not matter~\cite{zenke2021remarkable}. The slope $s$ of the derivative, however, does have a large influence on the training speed and final results. For this work, the slope $s$ has been set to 7.

Multiple challenges were observed during the training/deployment iterations. These are discussed here.
\subsubsection{Delay in SNN, training with time-shifted data}
\begin{figure}[h]
\centering
\includegraphics[width=0.48\textwidth]{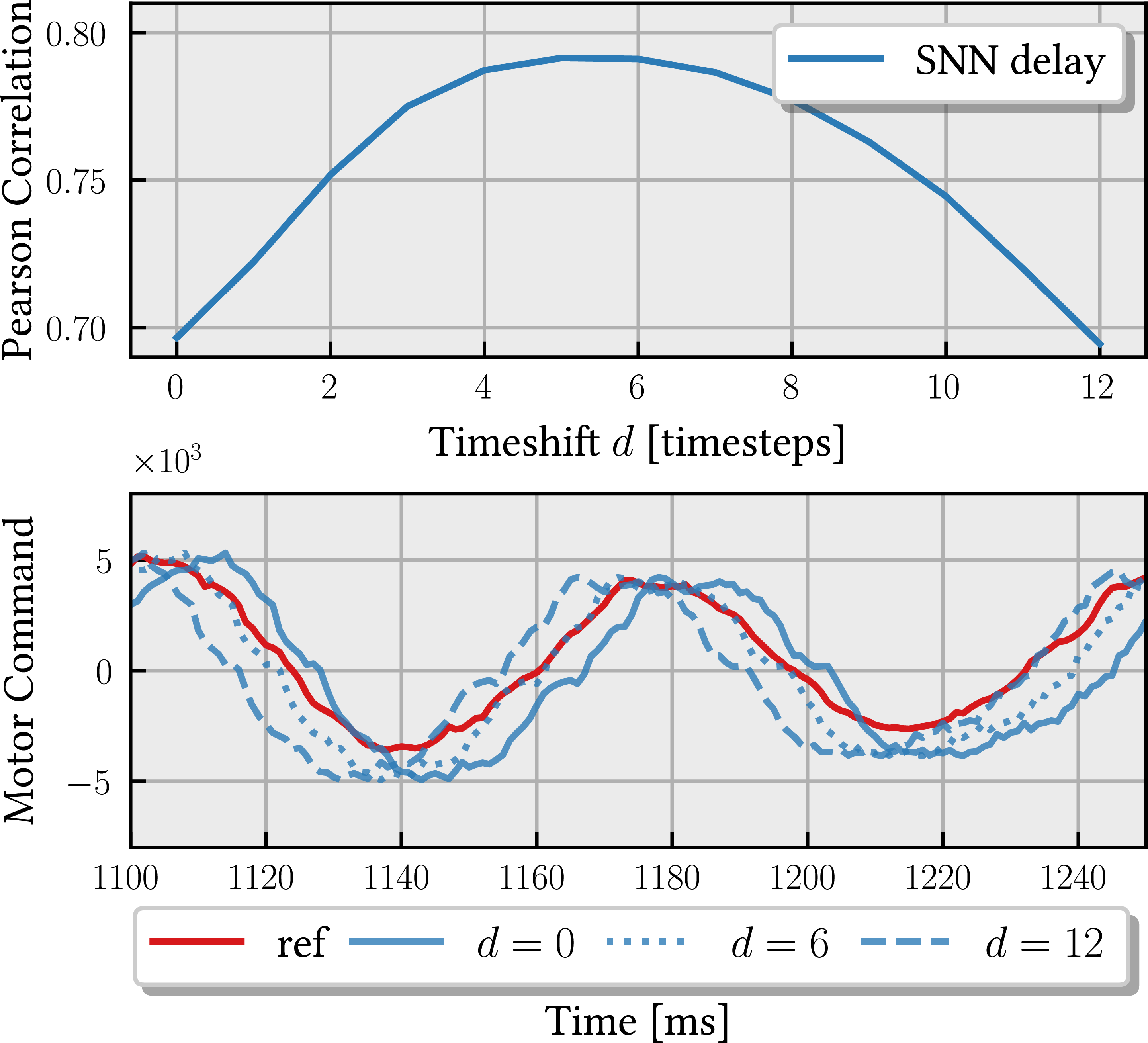}
\caption{Pearson Correlation between the output of the trained SNN and the regular PID output for different time shifts $d$. The bottom graph shows the output of the network for time shifts $d=0$, $d=6$ and $d=12$ compared to the target, further demonstrating that a delay is present in the network.  }
\label{fig:delay}
\vspace{-12pt}
\end{figure}

During training-implementation iterations, oscillations were observed on the real quadrotor. 
After investigation, these were attributed to a delay in the output of the network versus the target control signal. 
Due to the nature of the SNN with the implicit memory due to the leaking voltage and current, the output was delayed. This can be observed in Figure~\ref{fig:delay}. In the top part of the figure, the Pearson Correlation between the output of the SNN and the regular PID is compared for different shifts in time on the entire data set. In the bottom part of the figure, a small time sequence is shown that clearly shows the lag. The correlation is highest for 5-6 timesteps shift, indicating that this is indeed a problem when one trains SNNs for highly dynamic tasks that require a quick response to fast changes. In the case of a controller, a small delay in the derivative command will induce oscillations. 
To reduce this delay, and improve flight characteristics, we trained the control network on a time-shifted version of the target data. 
Specifically, we used the target signals of $\approx$~6 steps in the future. Consequently, the SNN needs to predict the reference control output in the future, which in turn results to less delay in the implemented controller.

\subsubsection{Imitation learning; reducing the reality gap}
The reality gap is a significant challenge in imitation learning particularly, since the reference controller only explores a limited portion of the state space around its stable behavior. This leads to a dataset that does not fully represent the full range of potential flight conditions or disturbances the SNN controller may encounter when deployed~\cite{ross2011reduction, ross2013learning}. Consequently, when the trained controller operates in real-world conditions, it can encounter "unseen" states or disturbances not present in the training data, resulting in unpredictable and unstable behavior. 

To address this, we expanded the training data to include a broader, more realistic range of states. Initially, the SNN controller was trained on data generated with the reference controller in the loop, as described in Section~\ref{subsec:training}. We then conducted additional data collection in two steps to diversify the training set: (1) flying the quadrotor with the initially trained SNN controller in the loop, while simultaneously logging the outputs the reference controller would have provided. This approach exposed the SNN to a set of states it is likely to encounter, fine-tuning the network around these points. (2) Introducing random disturbances to the regular PID controller's outputs to simulate unexpected environmental or system changes. Specifically, disturbances were applied to pitch, roll, and yaw commands at a 1\% probability per timestep (at 500Hz), lasting 0.2 seconds each, with disturbance size $X \sim U(0,50)\%$ of the absolute maximum command.

This additional data, including both the reference controller outputs and the effects of random disturbances, was incorporated into the training set. Retraining the SNN controller on this expanded dataset improved its robustness, enabling it to generalize across a wider range of states and disturbances, thereby reducing the likelihood of instability during real-world deployment.

\subsubsection{Splitting estimation and control}
As discussed in the section on architecture (see Figure~\ref{fig:fig1}), the network was split into an estimation and control part. 
If the network learning attitude estimation also has access to the control command, training is prone to end up at a local minimum. 
The network will then learn a function between control command and attitude; since the reference controller was in the loop this will be an easy function to learn. 
It can then completely disregard the sensor data, or only use it to slightly optimize the estimation. 
When this estimator is then used in the loop, the function between input command and attitude will be different since the trained controller is not perfect; this will further degrade flight performance. 
Hence, no connections between the input command and the attitude estimation layer are established.

\subsubsection{Integrator}
In developing an integrator within the spiking neural network (SNN) architecture, we faced challenges with parameter sensitivity, where small adjustments often led to significant errors or instability, causing the network to either underestimate the integral or diverge. This challenge is particularly acute in recurrent neural networks (RNNs), where recurrent gains above 1 often destabilize the system, while a recurrency lower than 1 produces a low-pass filter response. Orvieto~\textit{et al.}~\cite{orvieto2023resurrecting} have shown that carefully structuring RNN network architecture before training (e.g. by linearizing and diagonalizing the recurrency) is important to obtain the superior results of deep State Space Models (SSMs)~\cite{rangapuram2018deep}.

Another issue was the integrator signal’s dynamic: it shows large deviations at the start of a flight test but stabilizes quickly under constant disturbance. Effective integration through imitation learning required varying disturbances and resetting the initial integral for each sequence. Additionally, the integral signal changes more slowly than the proportional and derivative components, complicating the extraction of integral information from the total signal in a supervised-learning scheme.

\begin{figure}[h]
    \vspace{-8pt}
    \centering
    \includegraphics[width=\linewidth]{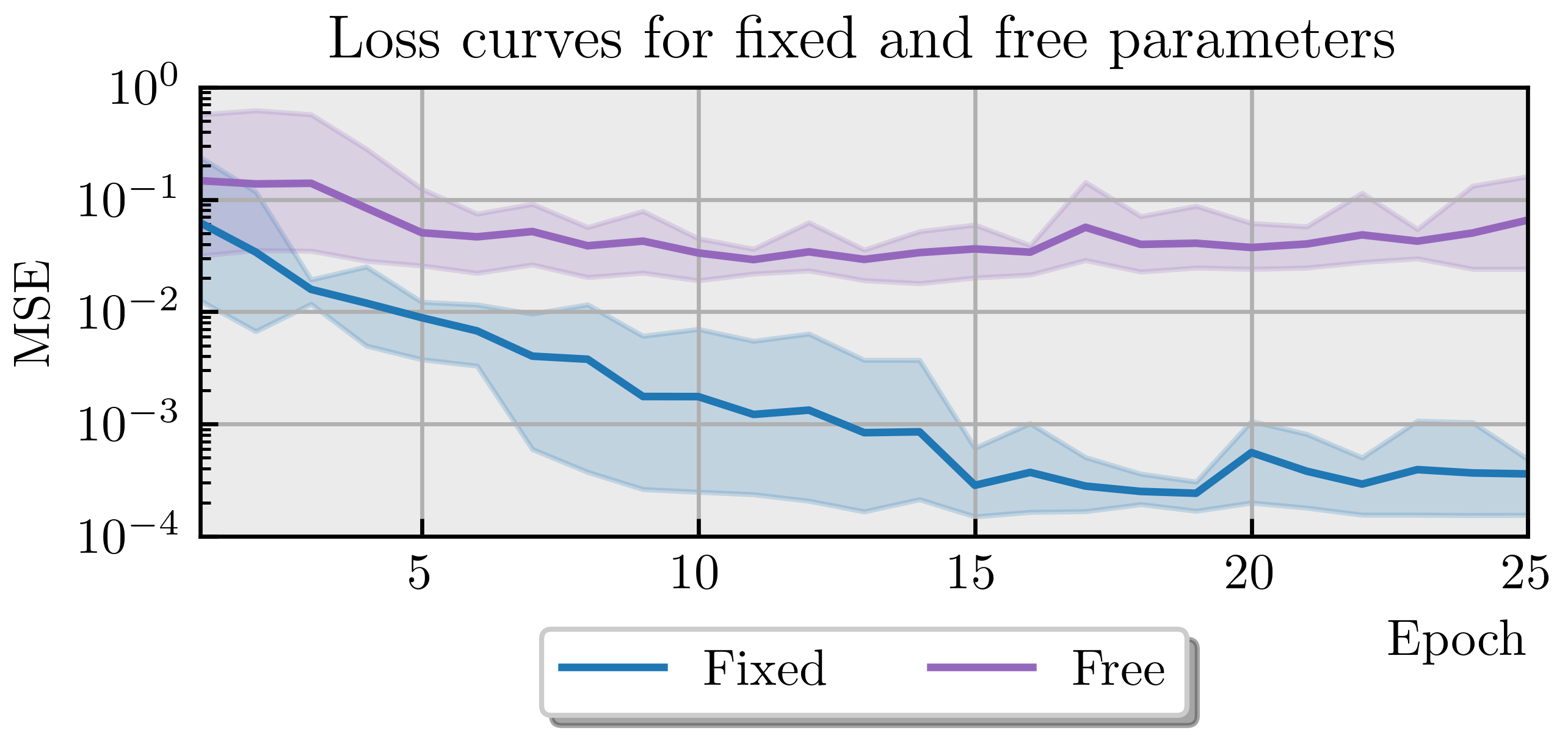}
    \caption{Training loss curves comparing fixed versus free neuron leak and threshold parameters. The proposed approach of fixing neuron parameters leads to stable convergence during training. Allowing these parameters to remain free results in training becoming trapped in local minima.}
    \label{fig:loss_curves}
    \vspace{-15pt}
\end{figure}

To address these issues for SNNs, we propose fixing certain neuron parameters within a small subgroup of neurons during training to ensure stability. 
Specifically, we set the leak parameters $\tau_i^\text{syn}$ and $\tau_i^\text{mem}$ and threshold $\theta_i$ of 10 neurons in the control layer to 1.
This allowed the neurons to integrate incoming signals without decay.
By training only the input and output weights and averaging spike outputs on integral data alone, we achieved a spike rate approximating the cumulative incoming signal, making the neuron responsive to transient and steady-state inputs. 
This approach is validated in Figure~\ref{fig:loss_curves}, which compares training curves for an integration task with fixed versus free neuron leak and threshold parameters. 
The fixed-parameter integrator provided the necessary stability, outperforming the fully unconstrained trained approach and satisfying the SNN-based system’s control requirements.
\subsection{Hardware setup}
To demonstrate the capabilities of our approach, we have implemented it in the control loop of the tiny open-source quadrotor Crazyflie~\cite{giernacki2017crazyflie}. 
By adding a Teensy 4.0 development board to the Crazyflie, the necessary computation power for running an SNN on a processor was obtained. 
This allowed us to run the complete SNN from input encoding to control commands at 500Hz in C++ on the ARM Cortex-M7 microprocessor.
To carry the extra weight of the Teensy, the regular 16mm brushed motors of the Crazyflie are swapped with 20mm brushed motors. 
To maximize the accuracy of the network while utilizing the Teensy to its full extent, the network was optimized for speed by removing unnecessary neurons. 
This was done by performing inference on a number of test sequences and calculating the total contribution of a neuron on the output by calculating the total number of spikes emitted multiplied by its weight to all outputs. 
Now the $N$ lowest contributing neurons can be removed from the implementation in C++ on the Teensy.
Although the network was trained with 150-150-130 neurons per layer respectively, we reduced the size to 150, 100, and 80 per layer respectively. By mainly pruning the neurons with recurrent connections this way, we almost halve the number of mathematical operations while retaining over 99\% of the original MSE that was used as the loss function during training. 

We send the attitude setpoints, along with the IMU measurements from the gyroscope and accelerometer, via UART to the Teensy deck. 
The neural controller's torque command outputs are transmitted back to the Crazyflie through the same UART connection, where they are incorporated into the motor mixer. The motor mixer is a linear transformation that converts torque commands into rotor velocities.
As the network runs at 500Hz in the loop, the maximum delay introduced in the system is 2 milliseconds. Even though this is fast enough to keep up with the lower-level control-loop in the Crazyflie, it might still influence the overall stability. 

An OptiTrack motion capture system provides accurate position measurement and an absolute heading. These are sent to the Crazyflie via a radio connection to a ground station laptop, which also handles the sending of high-level commands. 

The total take-off weight of the Crazyflie, including the Teensy 4.0 and upgraded motors, is only 35 grams. This allows for approximately 5 minutes of flight time.
\vspace{-5pt}
\section{Results}
\label{sec:results}
\vspace{-5pt}
\subsection{Position control}
\begin{figure}[h]
\centering
\includegraphics[width=0.5\textwidth]{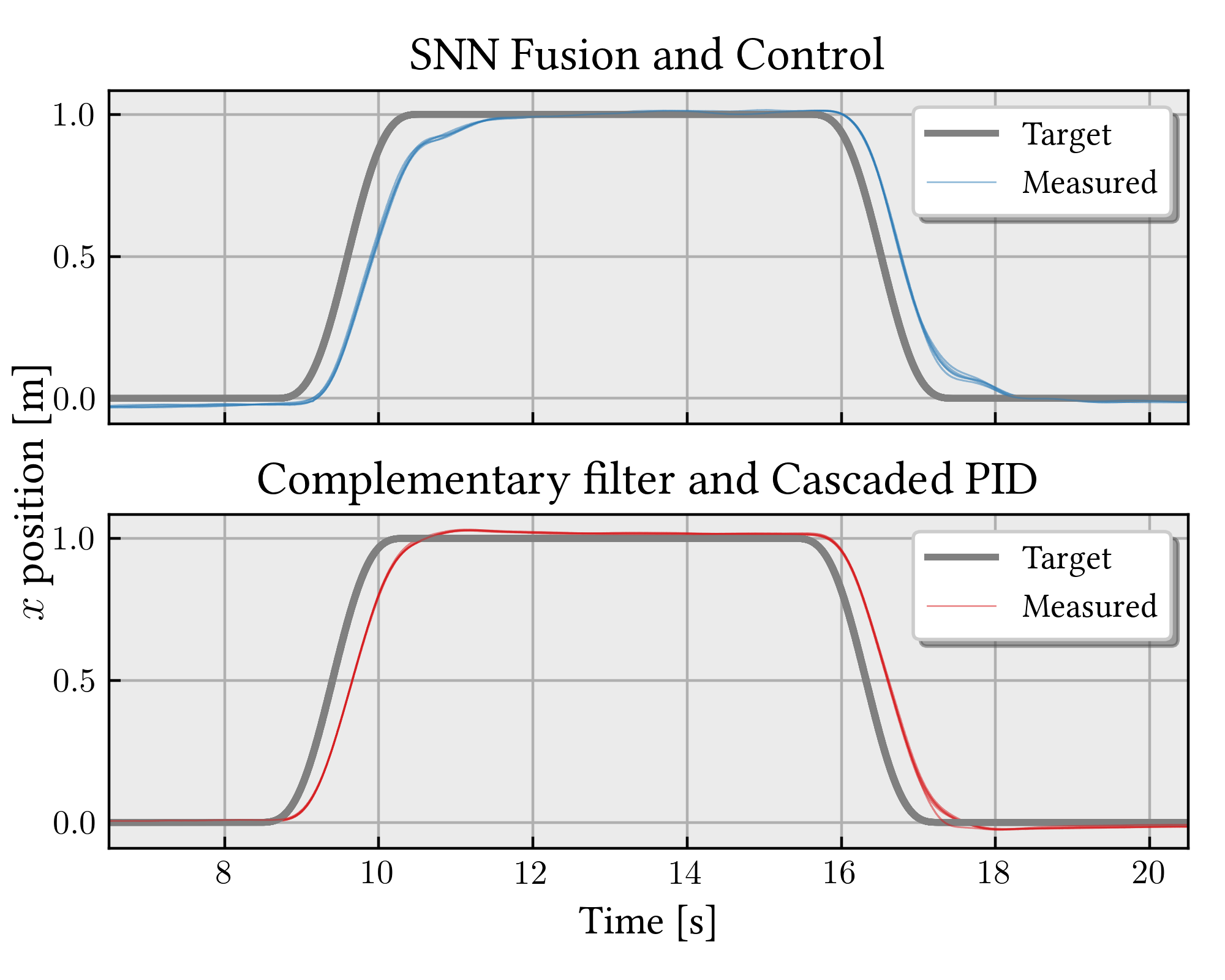}
\caption{Position step responses of the SNN system (top) and the regular PID flight stack (bottom) for 10 individual test runs. The SNN can accurately track the attitude references as given by the outer-loop position controller and maintain a stable flight path.}
\label{fig:pos_control}
\vspace{-15pt}
\end{figure}

To demonstrate the capabilities of the proposed SNN, we include it in a position control task. The higher-level attitude commands together with the IMU values are sent as inputs to the SNN, which produces pitch, roll and yaw torque commands. After a short period of hovering at $(x, y) = (0,0)$, the Crazyflie is commanded to move 1 meter in $x$-direction after which it is commanded to move back to $(0, 0)$. For both the SNN and PID controller, these tests were performed ten times. 
In Figure~\ref{fig:pos_control}, the position control results are shown. 
The results show that performing attitude estimation and control using an onboard SNN results in stable reference tracking, comparable to the regular flight stack of the Crazyflie.
\vspace{-8pt}
\subsection{Impact of Time-Shifted and Augmented Training Data on SNN Performance}
\begin{figure*}[!t]
    \centering
    \includegraphics[width=\textwidth]{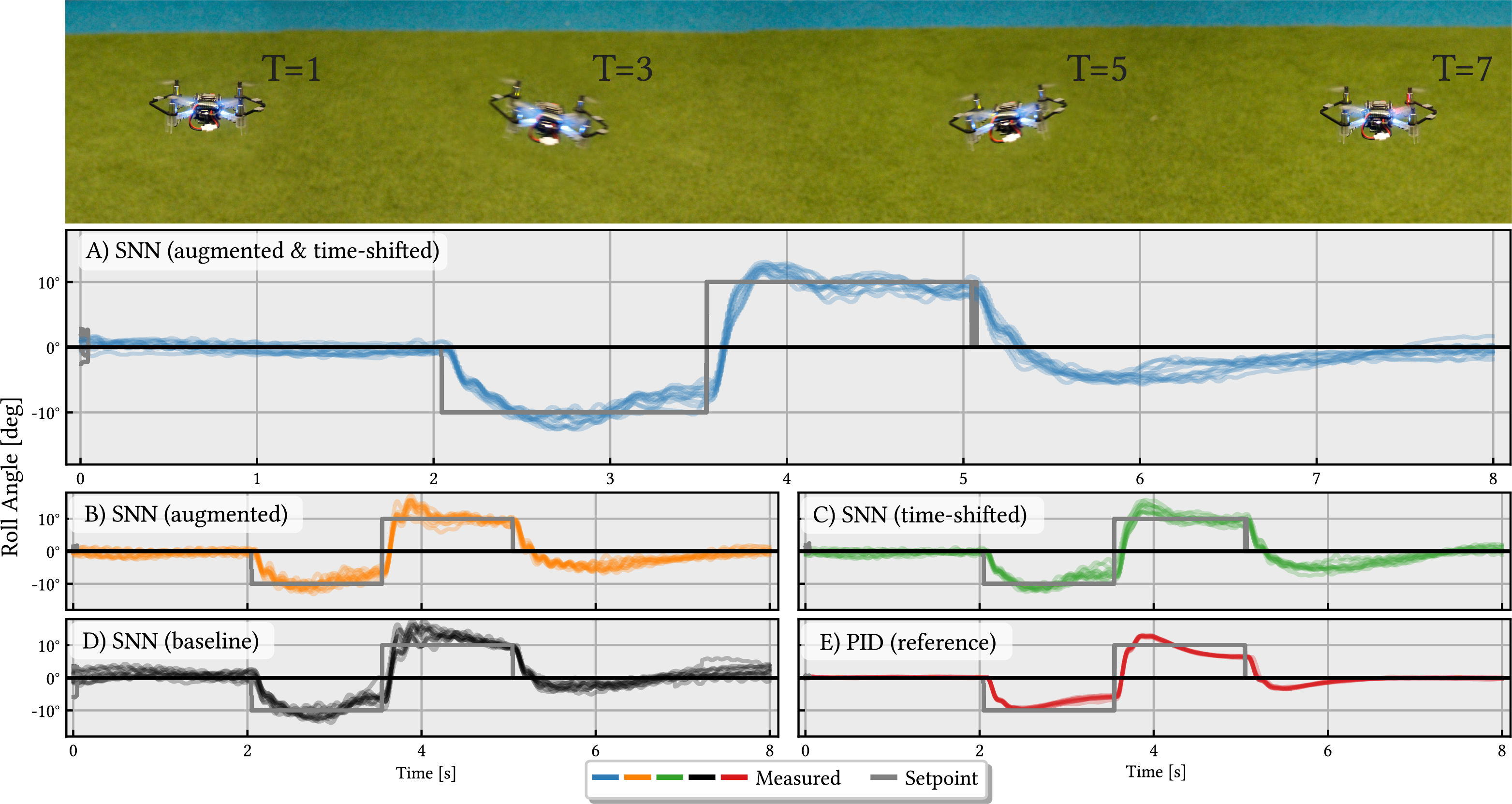}
    \caption{Attitude step responses of A) the fully-trained SNN system, B) the SNN trained with augmentation, C) the SNN trained with time-shifted data and D) the regular PID flight stack. The images on top show the Crazyflie during the different maneuvers.}
    \label{fig:att_control}
    \vspace{-13pt}
\end{figure*}

During testing, it was quickly identified that training the fusion network without augmenting the dataset does not produce a network that can be used in flight. Therefore, it was necessary to augment the dataset for this sub-network. However, to further investigate the behavior of the SNN and the influence of the modifications to the training procedure, another test is performed. Since the directly controlled variable is the attitude command, we compare the response of differently trained networks to an attitude setpoint change. For these tests, the Crazyflie received a roll setpoint of $0\degree$ for 2 seconds, followed by a setpoint of $+10\degree$ for 1.5 seconds, a setpoint of $-10\degree$ for 1.5 seconds before returning to a $0\degree$ setpoint for 2.5 seconds. Again, we performed ten tests per controller. The combined results of these ten tests per controller are shown in Figure~\ref{fig:att_control}, with A) the final SNN, B) the SNN that was trained on the augmented dataset, C) the SNN that was trained on time-shifted data, but without augmenting the dataset and D) the regular attitude estimator and controller on the Crazyflie.
The Root Mean Square Error (RMSE) between the commanded roll setpoint and the resulting (estimated) roll angle is given in Table~\ref{tab:rmse_comparison}, together with the average standard deviation (SD) of the response with respect to the average of all tests with the same controller. 
With a tracking error of only $3.03\degree$, the network is able to correctly estimate the attitude and also control it. 
Adding the suggested modifications to the training procedure reduces the tracking error from $3.24\degree$ to $3.03\degree$ compared to $2.67\degree$ for the reference controller (please note that the reference controller receives the estimated attitude directly, while the SNN needs to internally calculate this). 
Also, training on time-shifted data significantly reduces the oscillations as can be seen in Figure~\ref{fig:att_control}. 
This can also be inferred from the average SD that is significantly lower for the fully-trained SNN, showing that the controller performs more consistently across multiple tests. On the other hand, training on time-shifted data very slightly increases the rise-time (see Table~\ref{tab:rmse_comparison}). 
Since the increase is in the order of milli-seconds, it will not affect tasks like obstacle avoidance that generally operate in the 20-40Hz range~\cite{yu2023avoidbench} but it should be considered if it is used in super agile flight. 
\begin{table}[h]
\centering
\arrayrulecolor{black} % Make sure the table lines are black
\begin{tabular}{|>{\columncolor[gray]{0.9}}l|>{\columncolor[gray]{0.9}}c|>{\columncolor[gray]{0.9}}c|>{\columncolor[gray]{0.9}}c|}
\hline
\textbf{Controller}            & \textbf{RMSE} & \textbf{avg. SD} & \textbf{avg. RT}\\ \hhline{|=|=|=|=|}
SNN (time-shifted \& augm.)   & 3.03\degree & 0.77 & 145ms\\ 
SNN (augmented)         & 3.10\degree & 0.95 & 130ms\\ 
SNN (time-shifted)      & 3.24\degree & 0.92 & 145ms\\ 
SNN (baseline)          & 3.14\degree & 1.16 & 135ms\\
PID                     & 2.67\degree & 0.23 & 125ms\\ \hline
\end{tabular}
\vspace{0.1cm} 
\caption{Root Mean Square Error (RMSE), Standard Deviation (SD) and rise-time (RT) comparison between different controllers. Note that the PID receives the estimated attitude as input, while the SNN needs to calculate this internally.}
\label{tab:rmse_comparison}
\vspace{-20pt}
\end{table}
\vspace{-10pt}
\subsection{Power usage analysis}
The main benefits of having an end-to-end attitude SNN mainly derive from its combination with other autonomy functions such as computer vision on a single neuromorphic chip. 
Given the elementary nature of attitude estimation and control tasks, we do not expect any substantial performance or energy improvements for attitude estimation and control by itself. 

Still, we do think it is insightful to analyze the power usage of the current solution.
The SNN in this research runs on a conventional microprocessor, as currently available neuromorphic chips (like Intel's Loihi~\cite{davies2018loihi, orchard2021efficient} or SpiNNaker~\cite{furber2014spinnaker}) require supporting embedded systems that are too large for a 35-gram quadrotor or challenging to source. To explore potential power advantages, we performed some estimative calculations. Spike propagation through the network relies solely on additions rather than multiplications, allowing us to calculate the necessary operations based on addition alone. For the three-layer network used here, this would initially amount to approximately 42,500 additions per update. However, due to the 15-20\% sparsity in neuron activations at each timestep, the actual required operations reduce to around 7,500 additions.
In contrast, the cascaded PID controller on the Crazyflie requires about 28 additions and 52 multiplications per timestep. Moreover, a straightforward complementary attitude estimation filter will have as most expensive operation a non-linear \texttt{atan2} function that requires in the order of 15-30 multiplications. Since a 32-bit floating-point multiplication uses roughly 37 times more energy than a 32-bit integer addition~\cite{luo2024addition}, we can roughly equate the number of additions of a straightforward traditional pipeline with $\approx$~3,000 additions. Hence, on a conventional microcontroller, the SNN performs in the same energy order of magnitude as a PID-based controller. 

If small neuromorphic hardware becomes available that can natively support IMU readings, while implementing the SNN in hardware, energy consumption can be substantially reduced. Nonetheless, we maintain that the real gain would come when expanding this network to handle image data for instance, as seen in other neuromorphic works that show up to 100~$\times$ gains in efficiency (e.g.~\cite{paredes2023fully, orchard2021efficient}). This would create larger disparities due to the high multiplication demands in image processing tasks. 
Then, implementing all functionality in a single neuromorphic chip would make conventional companion computers obsolete, massively reducing energy consumption. 

Finally, further benefits can be expected when moving to event-based control, which has demonstrated potential for drastic reductions in computational load (up to 80\% for quadrotor attitude control~\cite{guerrero2014attitude}) by activating only when significant events occur. A drone in hover should only need to interfere and adapt its actuator commands when it starts to move, requiring no energy expenditure in between control events. Current microprocessors can not optimally benefit, because they still need to perform operations at a fixed frequency. 
\vspace{-8pt}
\section{Conclusion}
\label{sec:conclusion}
In this article, we have presented the first fully spiking attitude estimation and control pipeline for a quadrotor.
We show that by using imitation learning, it is possible to train a fully end-to-end SNN to control a micro drone. We augmented training data to further enhance the performance, using in-flight data. The network was also taught to predict a $k$-step advance control action to mitigate delays that are inherent to the SNN. These methods led to significant reductions in RMSE relative to the target attitude and decreased oscillations, collectively enhancing the drone's flight stability.
Furthermore, our findings indicate that constraining parameters during training to function as integrators improves training precision and information integration. For RNNs these parameters would be the recurrent weights, and for SNNs the leak and threshold parameters. This novel approach avoids local minima during training and allows for faster convergence. Next to that, our methods of implicitely learning integration and differentiation are not only applicable to attitude control for quadrotors, but apply to perception and control for robotics in general (e.g. using integration with rotary encoders or using differentiation to predict future states in model-based control).   
By evaluating the system’s performance in real-world conditions and comparing it with traditional control methods, we have laid the groundwork for future developments in neuromorphic control strategies.
The importance of a working imitation learning pipeline, for instance, has been demonstrated in~\cite{xing2024bootstrapping}, where the authors show that bootstrapping a RL pipeline with imitation learning results in more reliable RL training while outperforming imitation learning only. Our methods can thus be used to improve RL for SNNs. 

Future research should aim to implement these algorithms on neuromorphic hardware, which could yield substantial gains in energy efficiency and reduced latency, potentially extending flight times and enabling neuromorphic UAVs in energy-constrained scenarios. By advancing these techniques, we envision the next generation of highly efficient, adaptive, and intelligent UAVs.
\vspace{-8pt}
\section*{Supplementary Materials}
All code necessary to 1) train the SNN, 2) convert and run the SNN on a Teensy 4.0, 3) integrate in the Crazyflie firmware and 4) perform the tests can be found in \url{https://github.com/tudelft/neuromorphic_att_est_and_control}. The data that was used for training can be found here~\url{https://doi.org/10.4121/f474ef0a-6ef1-4ea1-a958-4827c4eadf60}.
\vspace{-8pt}

\ifCLASSOPTIONcaptionsoff
  \newpage
\fi

% references section
\bibliographystyle{IEEEtran}
\bibliography{bib}
\end{document}